\documentclass[11pt]{article}
\usepackage{fontawesome}

\usepackage[final]{acl}
\usepackage{times}
\usepackage{latexsym}
\usepackage[T1]{fontenc}
\usepackage[utf8]{inputenc}
\usepackage{microtype}
\usepackage{inconsolata}
\usepackage{graphicx}
\usepackage{booktabs}
\usepackage{tabularx}
\usepackage{array}
\usepackage{graphicx}
\usepackage{titling}
\usepackage{subcaption}
\usepackage{booktabs}
\usepackage{multirow}
\usepackage{tabularx}

\newcommand{\name}{ScheMatiQ}
\usepackage[normalem]{ulem}

\newcommand{\com}[1]{}

\newcommand{\resolved}[1]{}

\title{\texorpdfstring{
\raisebox{-0.35ex}{\includegraphics[height=1.5em]{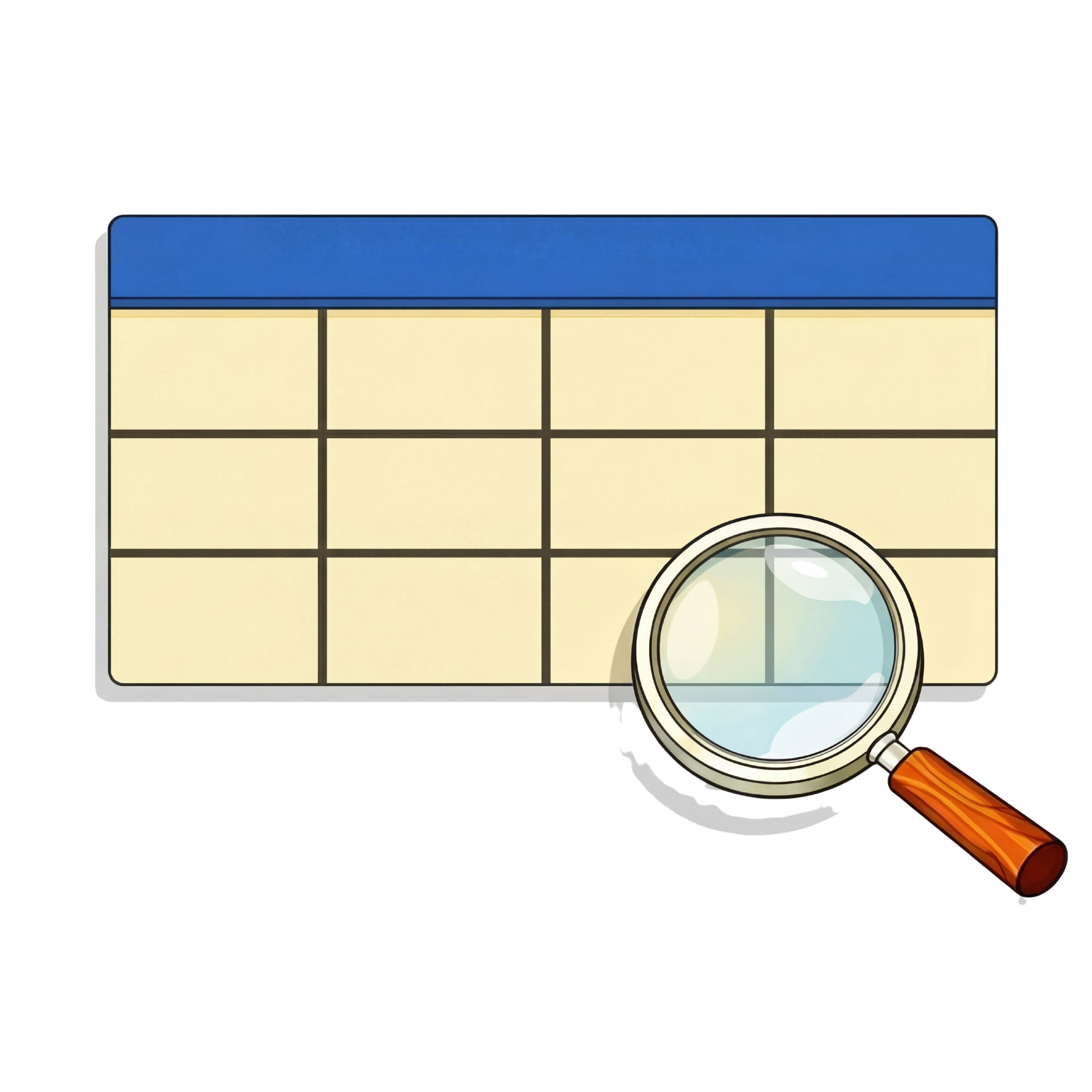}}
\hspace{0.1em}
\name{}: From Research Question\\
to Structured Data through Interactive Schema Discovery
}{\name{}: From Research Question to Structured Data through Interactive Schema Discovery}
}

\author{\\
\textbf{Shahar Levy}$^{1}$\thanks{~~Equal contribution.} \quad
\textbf{Eliya Habba}$^{1}$\footnotemark[1] \quad
\textbf{Reshef Mintz}$^{1}$ \quad \\
 \textbf{Barak Raveh}$^{1}$ \quad
\textbf{Renana Keydar}$^{2}$  \quad
\textbf{Gabriel Stanovsky}$^{1,3}$
\\
$^{1}$School of Computer Science and Engineering, The Hebrew University of Jerusalem \\
$^{2}$Faculty of Law, The Hebrew University of Jerusalem 
$^{3}$Allen Institute for AI
\\
{\{shahar.levy2, eliya.habba, gabriel.stanovsky\}@mail.huji.ac.il } 
\\
\faHome\ \href{https://www.schematiq-ai.com}
{\name{} Website} }

\begin{document}
\maketitle




\begin{abstract}
Many disciplines pose natural-language research questions over large document collections whose answers typically require structured evidence, traditionally obtained by manually designing an annotation schema and exhaustively labeling the corpus, a slow and error-prone process. We introduce \name{}, which leverages calls to a backbone LLM to take a question and a corpus to produce a schema and a grounded database, with a web interface that lets steer and revise the extraction. In collaboration with domain experts, we show that \name{} yields outputs that support real-world analysis in law and computational biology. We release \name{} as open source with a public web interface, and invite experts across disciplines to use it with their own data. All resources, including the website, source code, and demonstration video, are available at: \href{www.ScheMatiQ-ai.com}{www.ScheMatiQ-ai.com}.


\end{abstract}

\section{Introduction}
Across disciplines, research often begins with a natural-language question posed over a large collection of documents. For example, consider real-world questions from different fields: a legal scholar asking, \emph{Do judges appointed by different U.S. presidents differ in how they rule on immigration injunction cases?} in a large corpus of court decisions~\citep{klerman2025trump}; a computer scientist asking, \emph{When is Chain-of-thought (CoT) really helpful?} across hundreds of NLP papers~\citep{sprague2024cot}; or a computational biologist investigating whether \emph{It can be determined if a protein contains a nuclear export signal?} in a large collection of lab protocols~\cite{Xu2012_NESdb}.

Common to all such questions is the need to support answers with structured data over \emph{observation units}, the primary elements of interest implied by the research question and the corpus~\citep{blalock1960social}. For example, in the legal domain, this may be a Supreme Court justice.

Obtaining  structured data traditionally requires extensive manual effort across two mutually-informing stages. First, domain experts design an annotation schema that specifies the key question attributes (e.g., appointing president, ruling outcome) and potential confounders (e.g., age or education). Developing this schema requires domain knowledge and familiarity with the corpus. Second, annotators label the corpus according to the schema. This work, often delegated to research assistants, is expensive, slow, and vulnerable to human error~\citep{artstein-poesio-2008-survey}.

Though such research efforts are very common, they are not well supported by current LLM-based technologies, including many “deep research” solutions. These systems are typically geared toward retrieval rather than exhaustive processing, and they produce outputs that are difficult to interact with,  manipulate, or ground in the input texts.

\begin{figure*}[t]
  \centering
  \includegraphics[width=\linewidth]{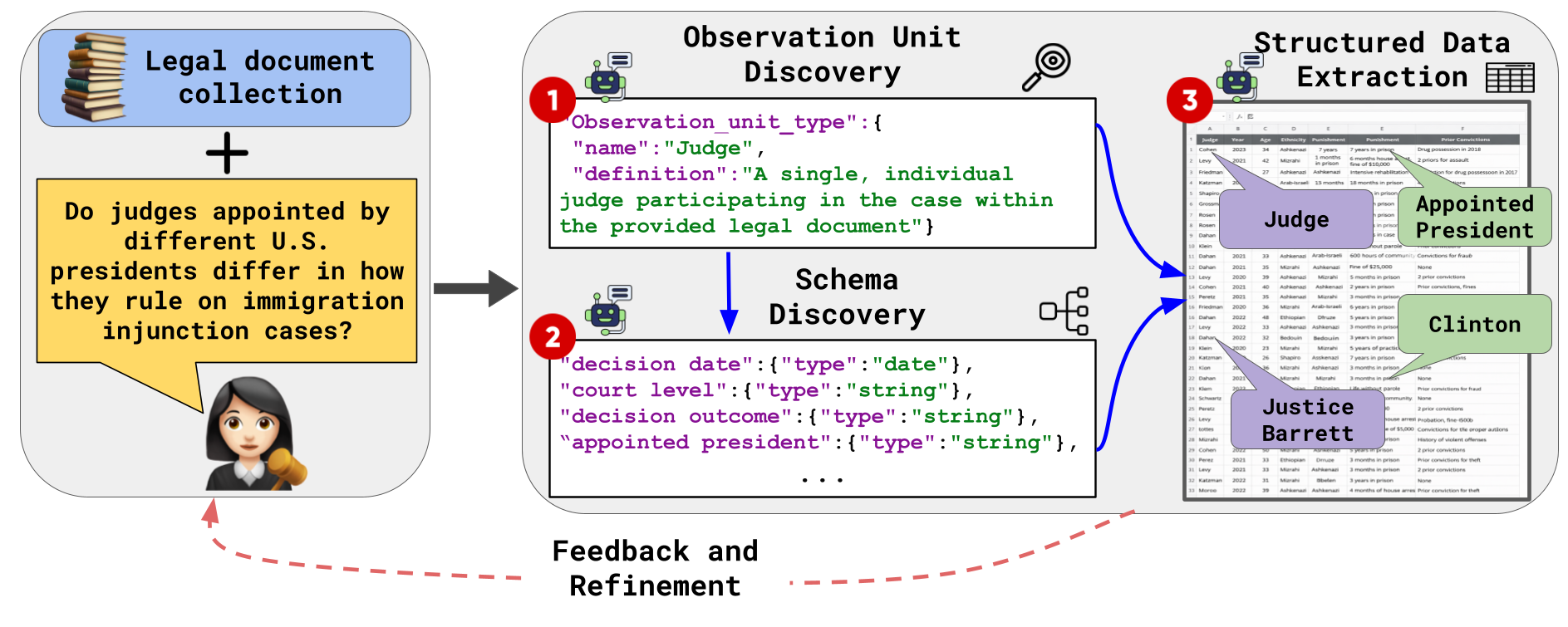}
  \caption{
    \textbf{\name{} workflow}. Given a natural-language question and a document collection, the system (1) discovers the observation unit, (2) discovers a query-guided schema, and (3) extracts structured values from the documents. Researchers can refine the schema and results through an interactive feedback loop.
    }
  \label{fig:pipeline}
\end{figure*}

In this work, we present \name{}, a framework that helps domain experts analyze large document collections around a guiding research question. As illustrated in Figure~\ref{fig:pipeline}, \name{} leverages calls to a backbone LLM to identify observation units, induce an annotation schema, and generate a structured database, grounding each output in the source documents so users can verify the evidence behind it. A dedicated user interface lets experts iteratively steer the extraction process by inspecting and revising schema elements.

We evaluate \name{} on two real-world use cases, in close collaboration with domain experts in law and computational biology. These settings pose distinct challenges: legal analysis often hinges on long-form arguments, whereas computational biology frequently demands numerical, protocol-grounded reasoning. In both settings, \name{} generates structured outputs that matches the vast majority of human-annotated schemas and introduce new attributes that experts find useful. 

We make \name{} fully open-source, and make it easy to use through a public web interface. We invite domain experts across disciplines to use it with their own questions and document collections, and NLP researchers to use it as a testbed for studying challenges such as long-context processing, efficiency, and effective user interfaces.

Our contributions are as follows:
(1) We introduce \name{}, a framework for automatic schema discovery and structured data extraction from an expert’s natural-language question and a collection of documents.
(2) We design and implement an interactive web-based system that supports human--AI collaboration.
 (3) We conduct an evaluation with domain experts in two real-world domains, showing \name{} recovers human-annotated schemas while also adding new valuable information.

\section{\name{} Principles}
\label{sec:principles}

We design \name{} around three core principles that reflect the real needs of experts in various disciplines.

\paragraph{Query-Driven Discovery.}
\name{} grounds the entire pipeline in the \emph{expert’s natural-language query}. 
We will show that different research questions over the same documents can lead to different observation units and, in turn, different data structures.

\paragraph{Human-in-the-Loop.}
\name{} keeps experts in control by making every component editable. Since experts bring essential domain knowledge, the system is designed to integrate their feedback at every stage. This principle ensures that the final dataset reflects both the model’s suggestions and the expert’s expertise.

\paragraph{Grounded and Traceable Outputs.}
\name{} grounds each of its outputs in the source documents. This allows experts to verify results, assess extraction quality, trace unexpected outputs, and ultimately trust that the final dataset is reliable and interpretable.

\section{\name{}}
\label{sec:system}

\name{} consists of three steps as illustrated in Figure~\ref{fig:pipeline}. First, given a natural language query and a collection of documents, the system \emph{discovers the observation unit}: the entity that each instance of the data should represent (Section~\ref{subsec:observation-units}). Second, using the documents, research question, and discovered observation unit, \name{} \emph{discovers the schema} by iteratively refining the list of fields relevant to answering the question as it processes the documents (Section~\ref{subsec:discover-schema}). Third, \name{} \emph{extracts values} for the fields in the discovered schema across all documents, producing an output structured database (Section~\ref{subsec:extract-table}). Throughout the process, experts can revise both the schema and the extracted data through human–AI collaboration. 

Below we elaborate on each of these steps, and provide prompt details in the Appendix~\ref{sec:app_prompts}.

\subsection{Observation Unit Discovery}

The first step is to identify the observation unit type, defining the structure of the resulting data by specifying what object each instance represents~\cite{blalock1960social}. 

For instance, in \emph{Do judges appointed by different U.S. presidents differ in how they rule on immigration injunction cases?}, the type of the observation unit is a Supreme Court justice. For \emph{When is Chain-of-thought helpful?}, the type is a single model evaluation under a specific experimental configuration. And for \emph{Can it be determined whether a protein contains a nuclear export signal?}, the type is an individual protein.

\label{subsec:observation-units}
\begin{figure}[htbp]
    \centering
    \begin{subfigure}{\columnwidth}
        \centering
        \caption{Diagram of the observation unit discovery flow.}
        \includegraphics[width=\linewidth]{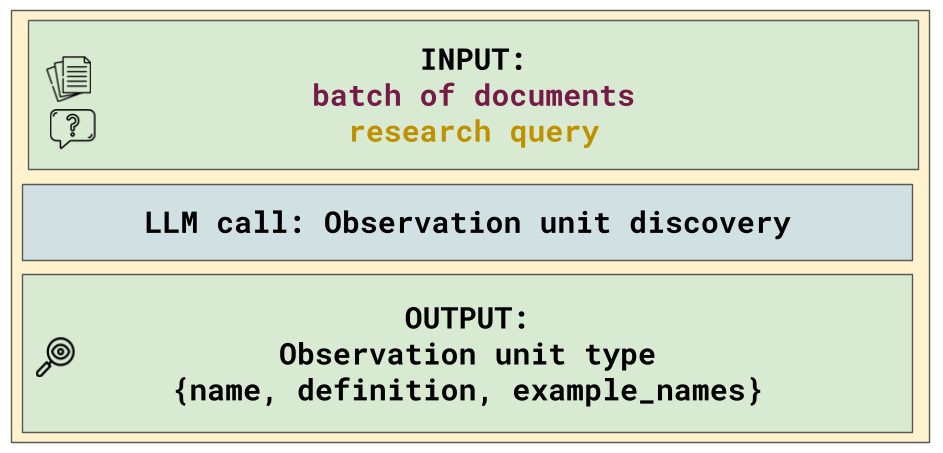}
        \label{fig:diagram-ou}
    \end{subfigure}
    \hfill
    \begin{subfigure}{\columnwidth}
        \centering
        \caption{Diagram of the schema discovery flow.}
        \includegraphics[width=\linewidth]{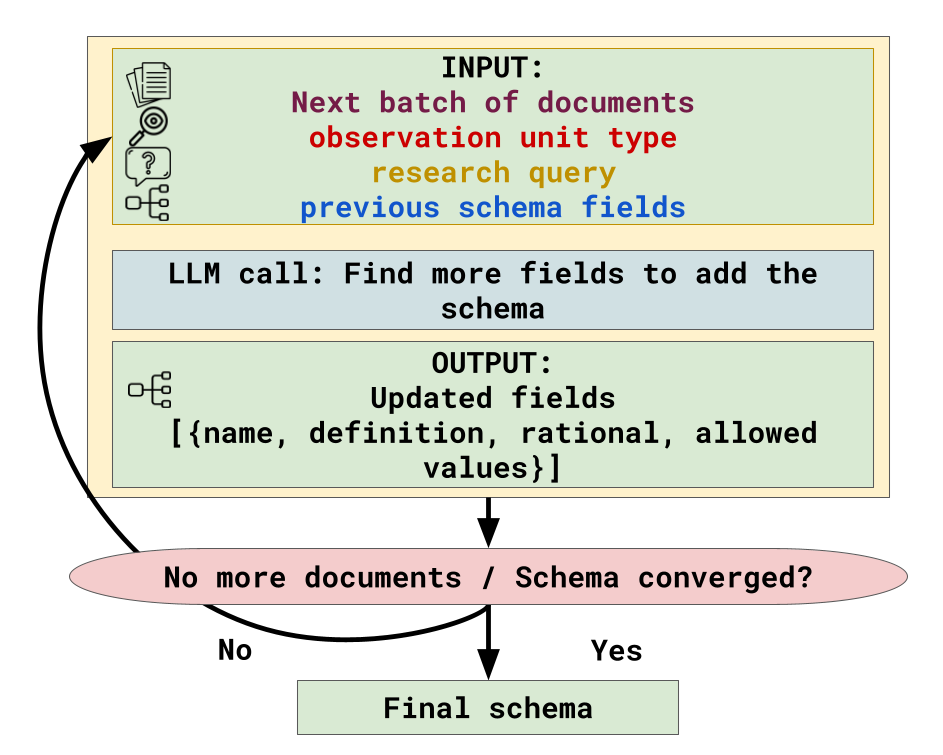}
        \label{fig:diagram-schema}
    \end{subfigure}
    \hfill
    \begin{subfigure}{\columnwidth}
        \centering
        \caption{Diagram of the structured data extraction flow.}
        \includegraphics[width=\linewidth]{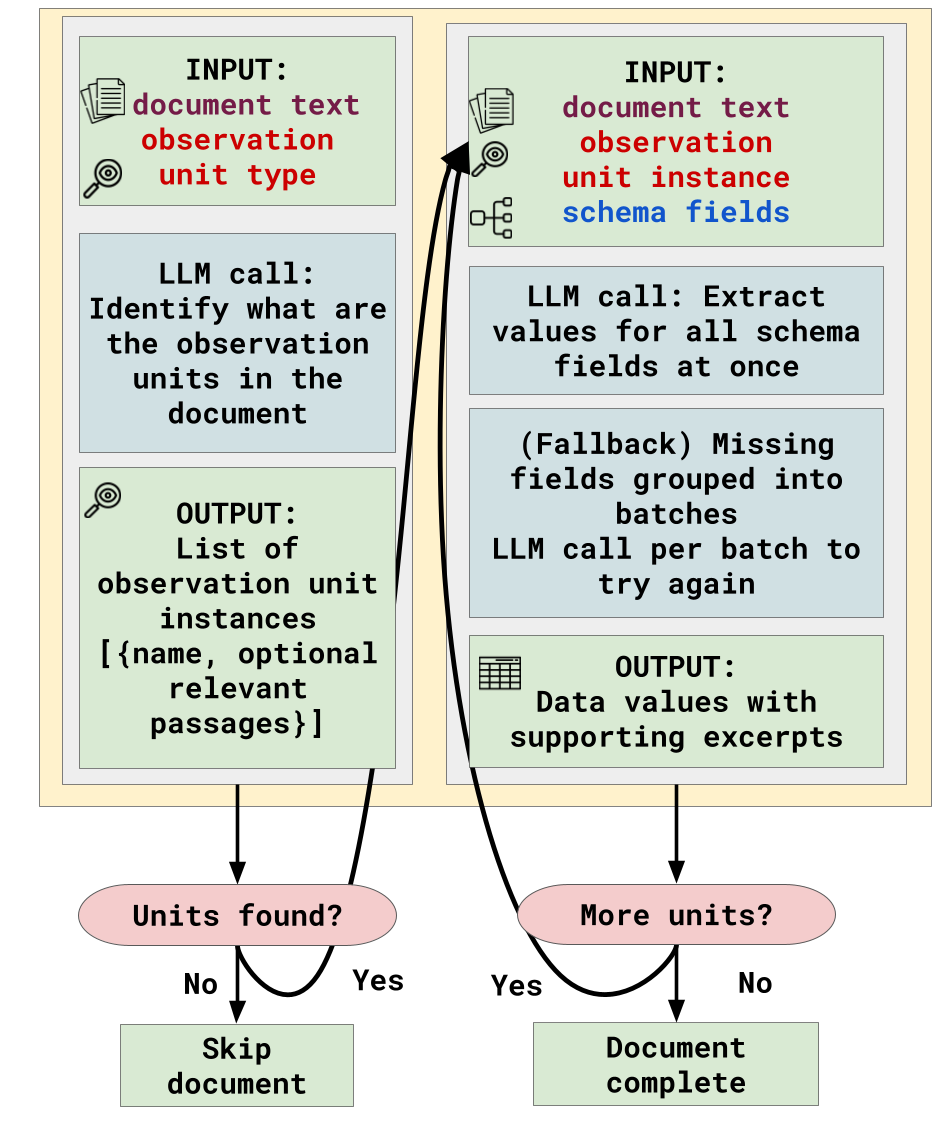}
        \label{fig:diagram-structured-data}
    \end{subfigure}

    \caption{
        Diagrams illustrating the three system components described in Section~\ref{sec:system}.
        Each panel shows the corresponding stage in the pipeline.
    }
    \label{fig:diagram}
\end{figure}

\begin{figure*}[t]
    \centering
    \includegraphics[width=\textwidth]
    {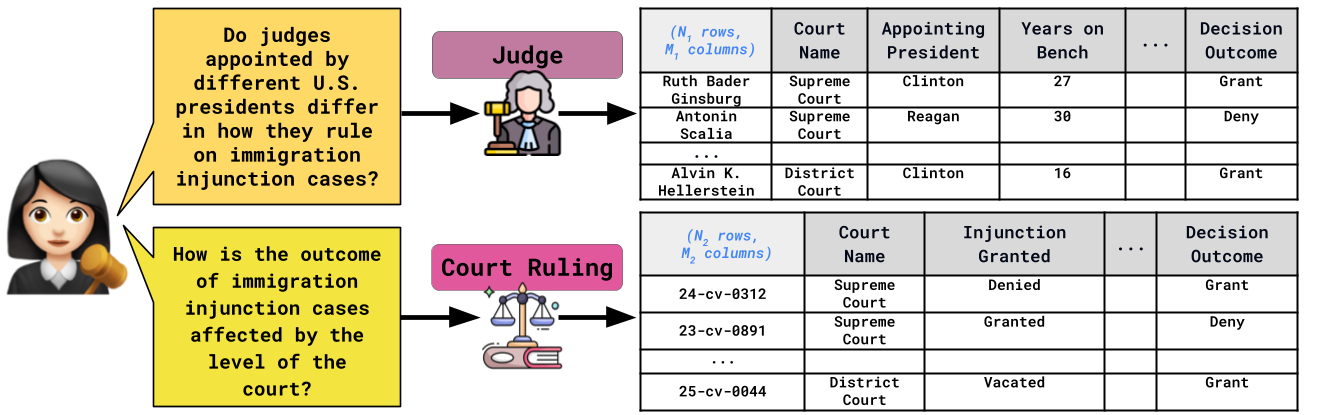}
    \caption{
    Different research questions over the same collection of documents lead to different observation units. 
    A judge-level question (top) yields one row per judge, while a case-level question (bottom) yields one row per court ruling, resulting in different schemas and table structures.}
    \label{fig:different_Q}
\end{figure*}

The relationship between documents and observation units is many-to-many: a single document may discuss multiple observation units, and the same observation unit may be discussed in multiple documents. Figure~\ref{fig:different_Q} illustrates how different research questions imply different observation units and, in turn, different data structures and document–observation-unit relationships.

\begin{figure*}[t]
  \centering
  \includegraphics[width=\linewidth]{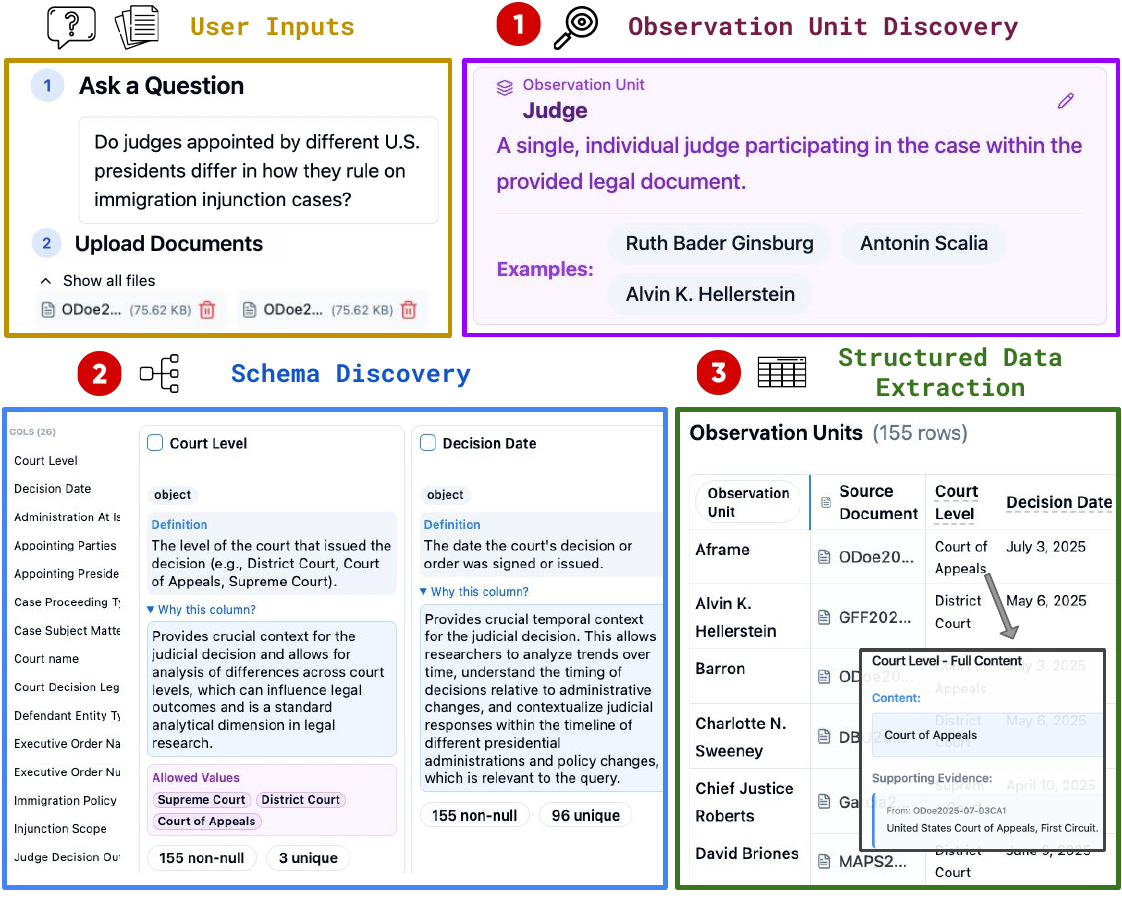}
  \caption{
    Screenshots of the \name{} web interface. Users provide a query and documents, inspect and refine the discovered observation unit and schema, and interact with the extracted table. }
  \label{fig:interface}
\end{figure*}

To identify the type of the observation unit, as illustrated in Figure~\ref{fig:diagram-ou}, we perform one LLM query using the expert’s research question together with a batch of documents, asking it to “identify what type the query is asking for.” The output of this step specifies the \emph{observation-unit type}, along with a \emph{description} of how it appears in the documents, and \emph{example instances} either from the input documents or from the model's parametric data. These outputs are displayed in the web interface, as shown in Figure~\ref{fig:interface}.

\paragraph{Human-in-the-Loop:}
Experts can revise the predicted type of observation unit or specify it manually if it is known in advance. This flexibility ensures that the resulting data will be structured around a desired entity.

\subsection{Schema Discovery}
\label{subsec:discover-schema}

After identifying the observation unit type, we discover \emph{the schema of the resulting data structure}: a set of  attributes that describe each observation unit (e.g., a particular judge, experiment, or protein) in ways that are relevant to answering the research question. For example, the schema in Figure~\ref{fig:pipeline} includes, for each Supreme Court justice, the appointing president and the outcome of their decision, among other relevant fields.

Designing the schema is a crucial step in answering research questions over document collections, and it is traditionally constrained by human capacity. If key factors are omitted, the analysis may miss important explanations or confounders. For example, a judge’s age or seniority could mediate decision-making, but would be invisible if not encoded in the schema. In manual workflows, schemas are typically shaped by the expert’s domain knowledge, preconceptions, and familiarity with the corpus. These limitations become especially acute for large collections.

\name{} enables a more accurate, scalable human-computer workflow by leveraging LLMs to surface important attributes \emph{across the entire document collection}. As illustrated in Figure~\ref{fig:diagram-schema}, we discover the schema by iteratively processing document batches and asking an LLM: “Do these documents suggest adding or refining the schema?”. The output specifies, for each field, a free-form \emph{definition} and a \emph{rationale} explaining how the field supports answering the research question, along with optional \emph{allowed values}, for example, whether the field should be numerical or free-form text. These outputs are displayed in the web interface, as shown in Figure~\ref{fig:interface}, and support human verification. They are also consumed by the data-extraction step. This process repeats until no new fields are proposed or the corpus is exhausted.


\paragraph{Human-in-the-Loop:}
\name{} supports two forms of schema intervention: (1) Field editing: modifying definitions or adding, removing, and merging fields; and (2) Incremental discovery: adding new documents after initial convergence, prompting the system to propose additional fields while preserving the existing schema.
These mechanisms enable iterative, flexible exploration as researchers expand their document collection and refine their understanding of the domain.

\subsection{Structured Data Extraction}
\label{subsec:extract-table}

Once the observation unit and full schema are obtained, we use them to annotate the document collection. The resulting structured data is represented as a table whose rows correspond to observation-unit instances and whose columns correspond to the schema attributes. This step mitigates the need for laborious and error-prone human annotation, enables downstream analysis of the extracted data, and it allows researchers to assess the schema by observing the values which populate each column and whether they capture meaningful patterns across the corpus.

As illustrated in Figure~\ref{fig:diagram-structured-data}, extraction is done in two stages. For each document, an LLM first identifies all instances of the observation unit (e.g., ``Ruth Bader Ginsburg'', or ``Antonin Scalia''). Then, for each instance, the LLM attempts to fill all schema fields in a single pass, and for any fields that remains unfilled, it performs a targeted follow-up extraction. All extraction is constrained by a strict evidence rule: a value can be extracted only if it is clearly supported by text in the document. Each output cell consists of  the \emph{extracted value} and the \emph{supporting evidence} grounded in specific text from the input documents, and displayed for experts in the web-interface as shown in Figure~\ref{fig:interface}. 

\paragraph{Human-in-the-Loop:}
Experts may correct or refine extracted cells, ensuring that the structured data reflects accurate, evidence-supported values. They can also add additional documents, allowing the table to expand as new data becomes available.



\section{System Evaluation}
\label{sec:evaluation}

Evaluating \name{} is challenging because it combines multiple components, human interaction, and large corpora of specialized texts, making direct end-to-end comparison to human annotation non-trivial. With domain experts, we study two use cases in empirical legal research and computational biology based on prior large-scale annotation projects, where a corpus, research question, schema, and human-annotated dataset already exist. This enables a direct comparison of \name{}’s outputs to human annotations, measuring agreement, omissions, and novel fields.

While these benchmarks are extremely challenging, reflecting several person-years of expert effort, they should not be treated as pure gold standard. Human-annotated schemas reflect feasibility constraints and can contain human errors. Ultimately, the value of \name{} is best measured by its real-world impact~\cite{reiter2025}, i.e., its adoption for \emph{new questions across disciplines}.


\subsection{Experimental Setup}
For each of our two domains, we specify the research question, the document corpus, and the human-annotated dataset. See additional implementation details in Appendix~\ref{sec:appendix_usecases}.

In all experiments we use the Gemini-2.5 family~\citep{comanici2025gemini25pushingfrontier}: Gemini-2.5-flash for observation-unit and schema discovery, and Gemini-2.5-flash-lite structured data extraction. 
The total cost of both of these uses cases is roughly 1 USD per 100 documents.

Users can specify other backbone LLMs by providing an API key to any model supported by \url{together.ai}.



\paragraph{Legal analysis.}
We follow \citet{klerman2025trump}'s analysis of 89 U.S. court decisions on immigration cases, asking \emph{Do judges appointed by different U.S. presidents differ in how they rule on immigration injunction cases?} To answer this, \citet{klerman2025trump} annotates each document with the judge name, appointing president, and decision outcome. 

\paragraph{Computational Biology.}
We use NESdb~\citep{Xu2012_NESdb}, a manually curated dataset of protein annotations in 96 scientific articles, asking if \emph{“it can be determined whether a protein contains a nuclear export signal? If so, how strong is it, and what is the confidence in that assessment?”}



\begin{figure}[t]
    \centering
    \includegraphics[width=\linewidth]{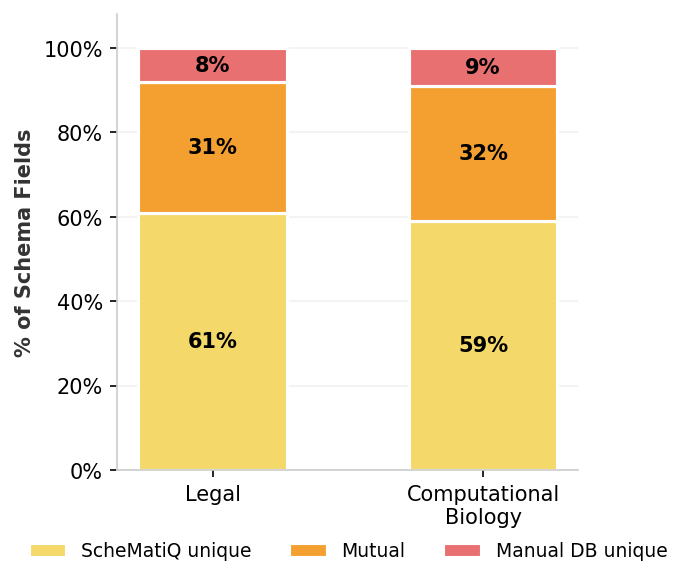}
        \caption{
        Schema-field coverage relative to manually curated gold schemas in the legal and computational biology domains. 
        Bars show the proportion of fields unique to \name{}, shared with the manual DB schema, or unique to the manual DB schema.
        }

    \label{fig:vennGold}
\end{figure}

\subsection{Results}
\label{sec:results}
Below we outline interesting conclusions derived from our experiments using \name{}:

\paragraph{\name{} successfully recovers gold schemas and contributes new, relevant fields.}
Domain experts first align the manually curated schema with the schema discovered by \name{},  then evaluate the fields that are unique to each schema. Figure~\ref{fig:vennGold} shows the resulting distribution of manual-only, shared, and \name{}-only fields across the two domains. In both settings, \name{} recovers all but two  broad miscellaneous fields. In contrast, the fields proposed by \name{} receive high relevance ratings, with mean scores of 4.2/5 in computational biology and 3.6/5 in the legal domain. 
For example, useful fields suggested in the legal domain include the legal basis for the court's decision, the scope of the injunction, and the presidential administration whose policy was challenged.

\paragraph{\name{}'s inputs are essential for capturing meaningful structure over real-world research questions.}
To assess the contribution of each input, we compare schemas generated under three configurations: using only the research question, using only the documents, and using both. Figure~\ref{fig:inputs} shows that question-only schemas tend to produce high-level, generic fields (e.g., Judge Name, Protein ID), while document-only schemas introduce broad content that is not necessarily aligned with the research question. In contrast, combining both inputs yields richer, context-specific fields (e.g., Immigration Policy Context, Mutation Description). The absence of a three-way overlap indicates that meaningful schemas do not emerge from either input alone; real-world research questions require query-dependent schema discovery.

\paragraph{\name{} successfully recovers observation units, while there's room for improvement for documents with many observations.}

In computational biology, \name{} identifies 87\% of proteins, and in the legal domain it identifies 97.5\% of the judges in the human-annotated dataset, with 82\% precision on the extracted set\footnote{The U.S. Supreme Court decides cases with all nine justices participating, and justices not named in an opinion are presumed to have joined the majority. The human-annotated dataset therefore records all nine justices for every Supreme Court decision. \name{}, by contrast, extracts only judges named in the document text. Accordingly, we do not count these unnamed justices as recall errors.}. This highlighting its potential to automate expensive annotation. Error analysis in both domains shows that most misses occur in documents containing many observation units, while recall is near-perfect when documents mention a single entity. Future work can specifically target these high-density documents.

\begin{table}[t]
\centering
\small
\begin{tabularx}{\columnwidth}{@{}lXrr@{}}
\toprule
Domain & Attribute & Precision & Recall \\
\midrule
\multirow{3}{*}{\raggedright\shortstack[l]{Computational\\Biology}}
& Export receptor  & 97.4\% & 90.7\% \\
& Detection method & 98.4\% & 75.9\% \\
& Source organism  & 70.9\% & 64.8\% \\
\midrule
\multirow{3}{*}{Legal}
& Court level    & 100.0\% & 100.0\% \\
& Decision date  & 97.2\%  & 97.0\%  \\
& Decision/Vote  & 99.2\%  & 91.6\% 

\\
\bottomrule
\end{tabularx}
\caption{\name{} value-extraction precision and recall by domain.}
\label{tab:cell_eval}
\end{table}

{\paragraph{\name{}’s value extraction is accurate but sensitive to normalization and missing evidence.}
In Table~\ref{tab:cell_eval}, we evaluate \name{}'s value extraction, by first aligning the observation units it identified with the original dataset units and then measure cell-level precision and recall. In each domain, we focus on attributes which can be evaluated reliably. 

In the computational biology domain, for the 162 proteins that were successfully aligned, two main patterns of errors emerge. First, \textit{naming granularity}: many errors in attributes like \textit{source organism} (60\% of errors) and \textit{export receptor} (13\% of errors) come from extracting surface forms (e.g., ''HPV-11,'' ''Mouse'') instead of the canonical names used in the human-annotated dataset (e.g., ''Human papillomavirus type 11,'' ''Mus musculus''). Second, the system shows consistently \textit{high precision}, indicating that when \name{} extracts a value, it is usually correct. Lower recall is often because \name{} omits values when there is no explicit evidence in the text, rather than making incorrect extractions.


In the legal domain, for the 143 judges that were successfully aligned, \name{} reliably extracts attributes stated explicitly in the documents: \textit{court level} with no errors and \textit{decision date} with 97\% accuracy. Most \textit{decision date} disagreements are attributable to the human-annotated dataset, including a typographical error (year ''3035'' instead of ''2025''). Errors concentrate on \textit{decision/vote}, where the label requires interpreting a judge's alignment with the outcome rather than reading a surface form. In most cases, \name{} abstains rather than predicts incorrectly, lowering recall (92\%) below precision (99\%).


\begin{figure}[t]
    \centering
    \begin{subfigure}{\linewidth}
        \centering
        \includegraphics[width=0.8\linewidth]{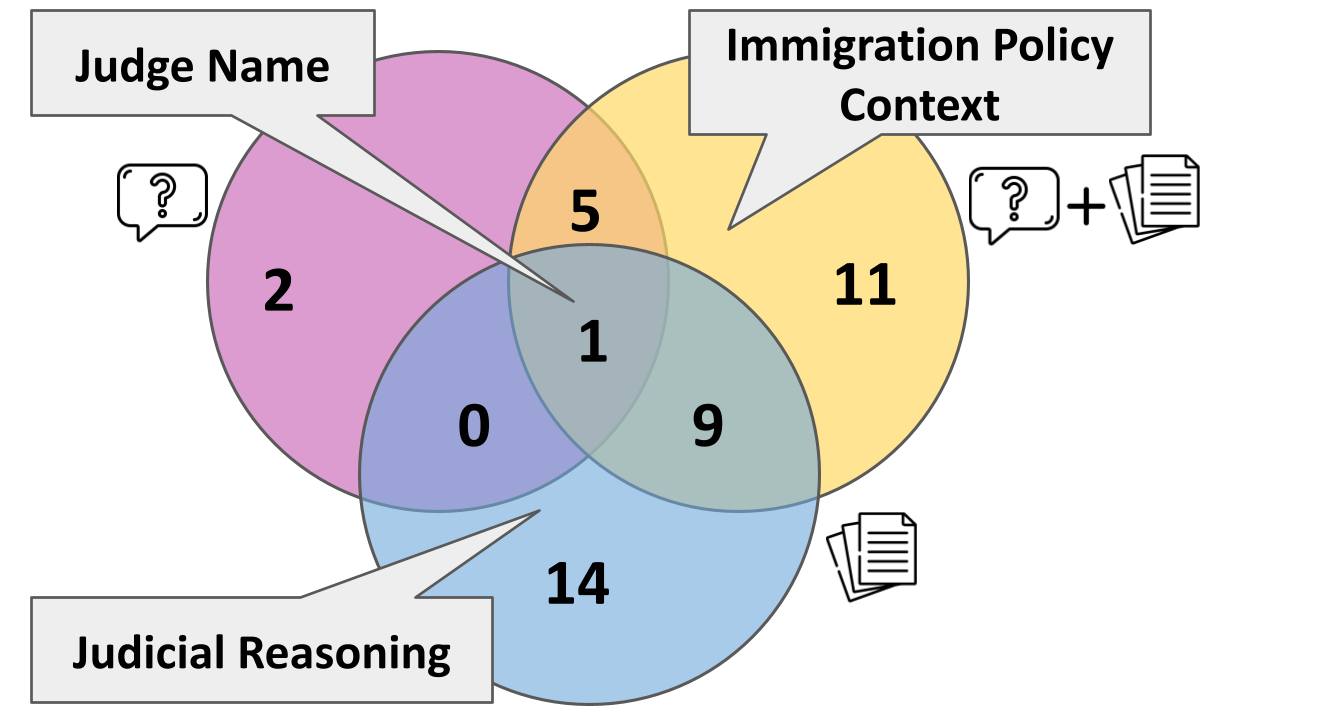}
        \caption{Legal domain.}
        \label{fig:inputs_legal}
    \end{subfigure}

    \begin{subfigure}{\linewidth}
        \centering
        \includegraphics[width=0.8\linewidth]{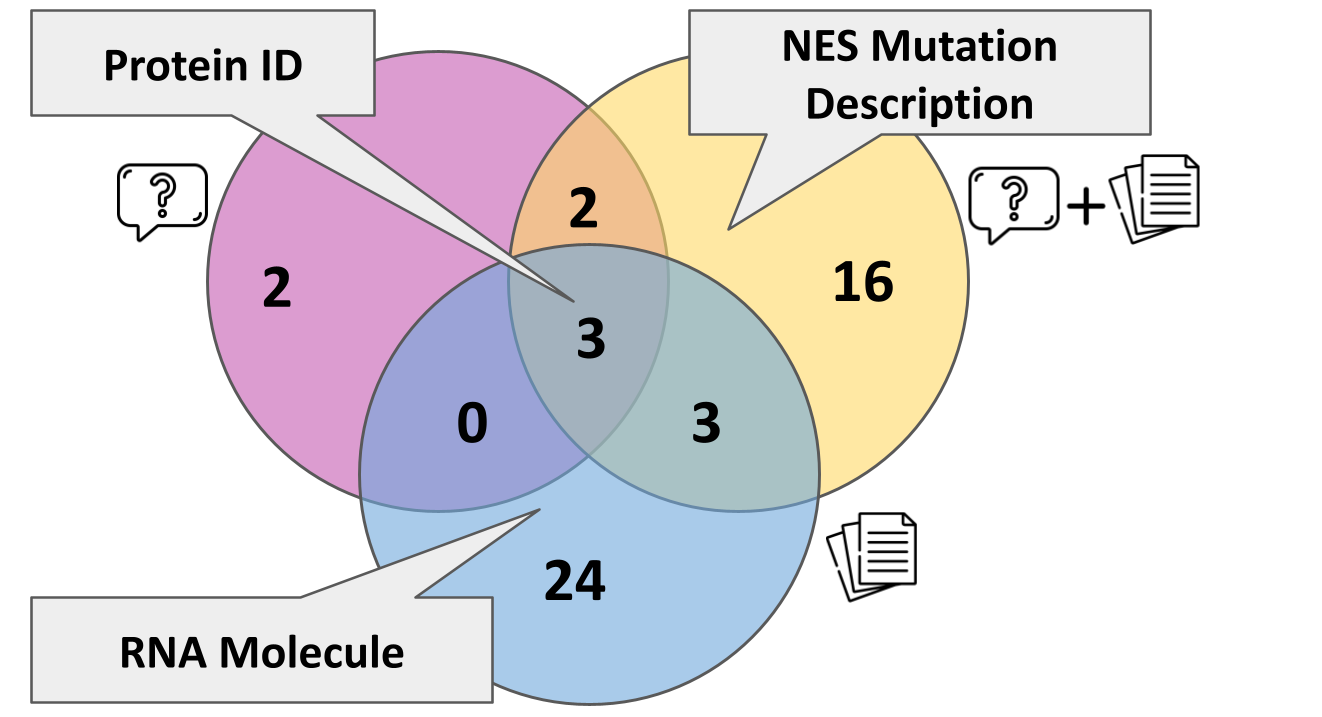}
        \caption{Computational biology domain.}
        \label{fig:inputs_bio}
    \end{subfigure}

    \caption{
    Schema-field overlap across three input conditions—query only (purple), documents only (blue), and the combined setting used by \name{} (yellow). 
    }
    \label{fig:inputs}
\end{figure}

\section{Related Work}
\label{sec:related_work}

Schema discovery from document collections has been studied across several
settings. Early text-to-table methods learn document-to-table mappings from
supervised pairs~\citep{wu-etal-2022-text-table}, without conditioning on a query. Later work generates
literature-review tables whose rows are papers and whose columns capture
aspects for comparison~\citep{newman-etal-2024-arxivdigestables}, separating schema
generation from value extraction, optionally
conditioned on a user-provided
intent~\citep{padmakumar-etal-2025-intent}. In all cases, rows remain
paper-level. Other systems support query-driven extraction:
\citet{jiao-etal-2023-instruct} produce a tabular output from a single text
without defining row-level structure across documents;
\textsc{SciDaSynth}~\citep{Wang_2025} generates tables from user questions,
with columns driven by the questions rather than the corpus; and
\textsc{schema-miner}~\citep{sadruddin2025llms4schemadiscoveryhumanintheloopworkflowscientific}
mines schemas from a domain specification, targeting reusable ontologies
rather than a specific research question.
\citet{dunn2022structuredinformationextractioncomplex} fine-tune LLMs over
scientific text for structured extraction but assume a predefined schema, and even with a fixed schema
\citet{ghosh-etal-2024-toward} find ad-hoc scientific extraction unreliable
without close expert review.

\name{} conditions schema discovery on \emph{both} the research question and
the documents, and explicitly identifies the \emph{observation unit} that
defines each row. Neither input alone yields context-specific schemas, and
experts can revise the unit, schema, and extracted values at any stage, so
\name{} supports specific research questions rather than generic document
comparison.
\section{Conclusion}
We introduced \name{}, an interactive framework for query-driven schema discovery and dataset construction. Given a research question and a corpus, \name{} identifies the appropriate observation unit, induces a question-specific schema, and extracts a structured dataset that experts can iteratively refine. Across empirical legal research and computational biology, our evaluation shows that \name{} produces meaningful schemas and supports practical research workflows.


\section{Limitations and Ethical Concerns}



Our experiments rely on closed-source LLM APIs, which makes full reproducibility difficult to guarantee. We observe small variations between runs even with fixed parameters, likely due to non-deterministic decoding or unannounced model updates by the provider. While these differences are typically minor, they may lead to slight changes in column naming or value extraction across runs. \name{} supports using open-weight models hosted locally, which can mitigate this issue.

\paragraph{Data privacy.} Users can opt in to have their data recorded for research purposes, otherwise, we do not store any session data.
\section{Acknowledgments}
This research was supported in part by Google.org and the Google Cloud Research Credits program for the Gemini Academic Program. 
We are grateful to Hadar Franco (\href{https://www.anicca-ai.com/}{Anicca.AI})  for her valuable help with the web interface and deployment.

\bibliography{custom}
\clearpage

\appendix
\section{Use Cases: Full Specifications}
\label{sec:appendix_usecases}

\paragraph{Legal Domain}

\textbf{Dataset.}
Court decisions of U.S. court cases concerning immigration policies and injunction proceedings.

\textbf{Full Query.}
Do federal judges appointed by different Presidents (Trump vs.\ other Republican vs.\ Democratic) differ in their voting tendencies on immigration injunction cases? Do Trump-appointed judges tend to be more supportive of Trump administration immigration policies compared to judges appointed by other Republican or Democratic presidents?

\textbf{Observation Unit — Judge.}
A single, individual judge participating in the case. If a case includes multiple judges (e.g., a panel), each judge is treated as a separate observation (row).

\textbf{Full Schema (Columns).}
Judges On Panel; Appointing Presidents On Panel; Appointing Parties On Panel; Policy Instrument Purpose; Plaintiff Immigration Status Type; Policy Instrument Type; Policy Instrument Issuing Authority; Court Decision Legal Basis; Decision Date; Immigration Policy At Issue; Executive Order Name; Legal Challenge Grounds; Defendant Entity Types; Injunction Scope; Policy Instrument Date; Judge Names; Judge Decision Outcome; Case Subject Matter; Administration At Issue; Policy Instrument Target Group; Executive Order Number; Judge Decision Tendency; Court Level; Case Proceeding Type; Plaintiff Entity Types; Court Name.

\vspace{1em}

\paragraph{Computational Biology Domain}

\textbf{Dataset.}
A collection of 110 scientific papers describing experimental studies of Nuclear Export Signals (NES) in proteins. The papers correspond to references scraped from NESdb.

\textbf{Full Query.}
Given a protein sequence, can it be determined whether or not it contains a nuclear export signal (NES)? If it does, how strong is the NES, and what is the confidence in that assessment?

\textbf{Observation Unit — Protein.}
A single protein or polypeptide sequence evaluated for the presence, strength, or characteristics of a Nuclear Export Signal (NES).

\textbf{Full Schema (Columns).}
NES Motif Count; Export Mechanism Type; NES Critical Residues; NES Presence Status; NES Activation Conditions; Regulatory Interacting Protein; NES Determination Evidence; NES Binding Affinity; NES Origin; NES Masking Agent; Competing Localization Signals; Export Receptor; NES Residue Coordinates; NES Identifier; NES Functional Impact; NES Transferability; NES Consensus Conformity; NES Strength Characterization; Protein Name; Reclassification Status; Source Organism; NES Conservation Status; Observed Subcellular Localization; NES Regulation Mechanism; NES Structural Domain; Identified NES Sequence.

\section{System Architecture}
\label{subsec:system-architecture}

The architecture of \name{} is organized into three main layers:

\begin{itemize}
\item \textbf{Frontend:} A React application built with TypeScript and Tailwind CSS. It provides an interactive interface for configuring queries and uploading input documents, editing schemas, and exploring extracted tables, with real-time updates streamed from the backend.

\item \textbf{Backend:} A FastAPI server that exposes REST endpoints for all pipeline operations. It also maintains a WebSocket channel to stream live progress updates (e.g., step-by-step extraction results) to the frontend.

\item \textbf{Core Library:} A standalone Python package implementing the core \name{} components: observation-unit discovery, schema discovery, and value extraction. The library supports multiple LLM providers, including OpenAI’s GPT-4~\cite{openai2023gpt4}, Google’s Gemini family~\cite{gemini2023capable}, and Together AI\footnote{\href{https://www.together.ai/}{https://www.together.ai}}  models. For local deployments, it also supports open-weight models hosted through the HuggingFace Transformers library~\cite{wolf-etal-2020-transformers}.
\end{itemize}

This separation enables researchers to use the core algorithms programmatically through the \name{} Python package, while the web interface layers add session management, cloud storage (Supabase), and an interactive human-in-the-loop editing flow. The entire system is deployed on Railway using Docker containers for portability and scalability.

\section{Prompt Templates}
\label{sec:app_prompts}
In this section, we present the core prompt structures guiding the \name{} discovery pipeline in Figure~\ref{fig:appendix:prompts}.

\begin{figure*}[t]
    \centering

    \begin{subfigure}{\textwidth}
        \centering
        \includegraphics[width=\linewidth]{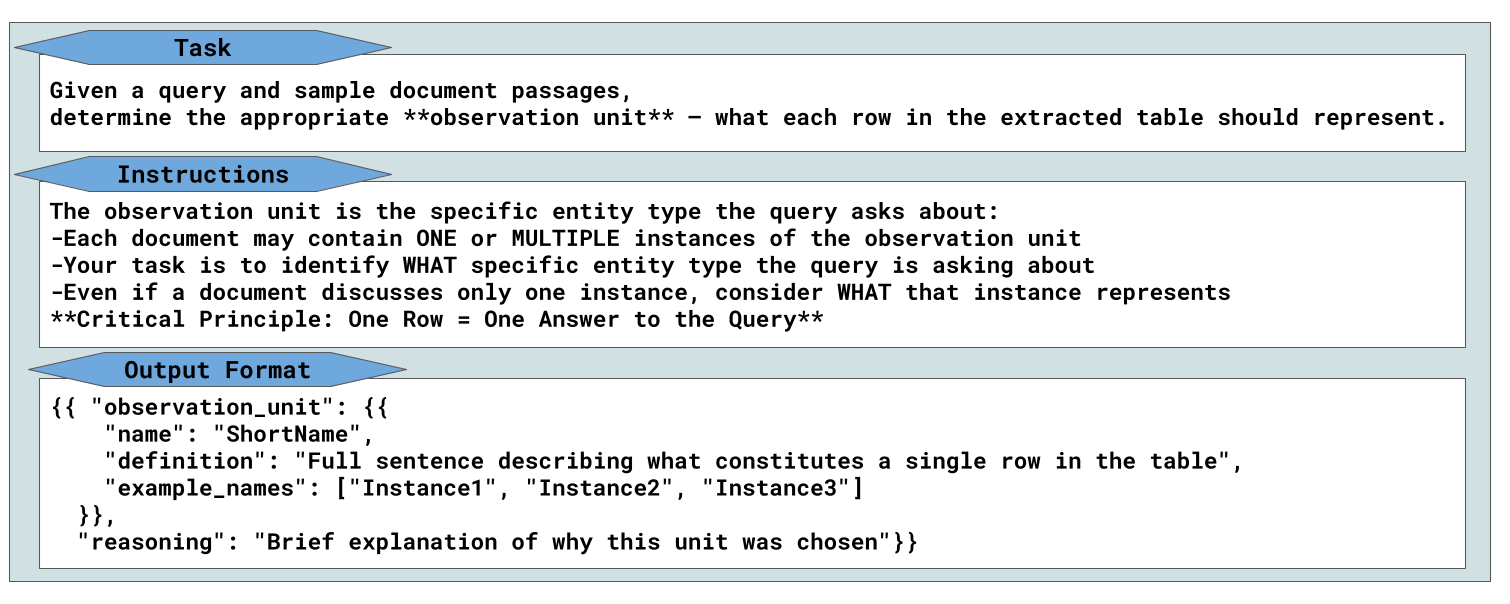}
        \caption{Simplified prompt for observation unit discovery.}
        \label{fig:appendix:prompts-ou}
    \end{subfigure}

    \begin{subfigure}{\textwidth}
        \centering
        \includegraphics[width=\linewidth]{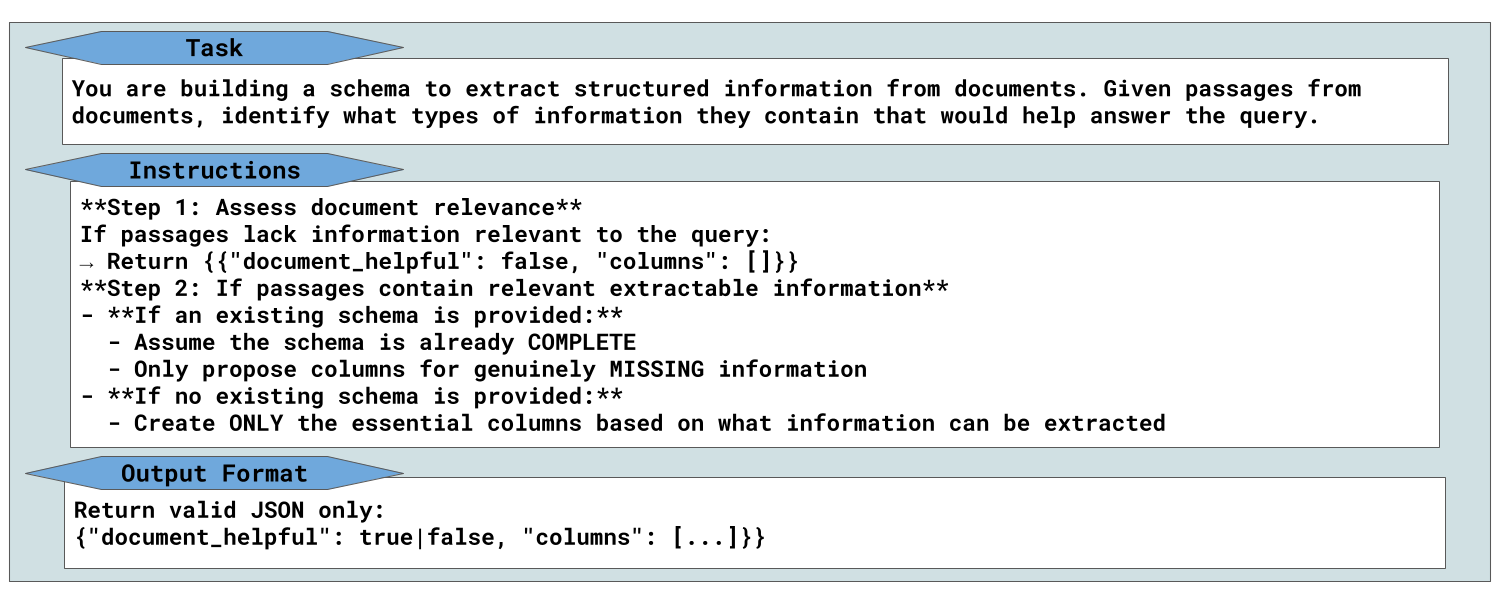}
        \caption{Simplified prompt for schema discovery.}
        \label{fig:appendix:prompts-schema}
    \end{subfigure}

    \caption{
        Simplified LLM prompt excerpts illustrating two core stages of the system pipeline:  
        (a) observation unit discovery and (b) schema discovery.  
        Full implementation details and complete prompts are available in our GitHub repository.
    }
    \label{fig:appendix:prompts}
\end{figure*}
\end{document}